\documentclass[conference]{IEEEtran}
\IEEEoverridecommandlockouts
\usepackage{cite}
\usepackage{amsmath,amssymb,amsfonts}
\usepackage{algorithmic}
\usepackage{graphicx}
\usepackage{textcomp}
\usepackage{xcolor}
\def\BibTeX{{\rm B\kern-.05em{\sc i\kern-.025em b}\kern-.08em
    T\kern-.1667em\lower.7ex\hbox{E}\kern-.125emX}}
\begin{document}

\title{Image Compression Using Singular Value Decomposition\\}

\author{\IEEEauthorblockN{Justin Jiang}
\IEEEauthorblockA{School of Computer Science \\
Carnegie Mellon University \\
Pittsburgh, United States \\
Email: justinji@andrew.cmu.edu \\
ORCID: 0009-0005-9513-1680}
}

\maketitle

\begin{abstract}
Images are a substantial portion of the internet, making efficient compression important for reducing storage and bandwidth demands. This study investigates the use of Singular Value Decomposition and low-rank matrix approximations for image compression, evaluating performance using relative Frobenius error and compression ratio. The approach is applied to both grayscale and multichannel images to assess its generality. Results show that the low-rank approximations often produce images that appear visually similar to the originals, but the compression efficiency remains consistently worse than established formats such as JPEG, JPEG2000, and WEBP at comparable error levels. At low tolerated error levels, the compressed representation produced by Singular Value Decomposition can even exceed the size of the original image, indicating that this method is not competitive with industry-standard codecs for practical image compression.
\end{abstract}


\section{Introduction}
Images are a substantial portion of internet traffic, and their storage and transmission place increasing demands on modern systems. Reducing the number of bytes required to represent an image is important for both storage efficiency and bandwidth usage. This study examines using Singular Value Decomposition to produce low rank approximations which retain most visual details while reducing storage.

To evaluate the quality of these approximations, this paper uses Frobenius error to measure the difference between an approximation and the original image. Because absolute Frobenius error scales with image size, relative Frobenius error is used to compare performance across images with different resolutions. Compression ratio is also measured to quantify the amount of storage saved.

The evaluation uses a 1,000 image subset of the ImageNet Large Scale Visual Recognition Challenge 2017 test set, which contains a wide range of objects, scenes, and lighting conditions. For each image, target relative Frobenius error levels from 0.05 to 0.95 are tested to determine the minimum rank required to satisfy each tolerance. This produces a tolerance curve that shows how compression ratio changes as the allowed error increases. The results indicate that compression improves rapidly for moderate error tolerances but becomes terrible when very low error is required.

To contextualize the efficiency of compression using Singular Value Decomposition, the same images are also compressed using JPEG\cite{wallace1992jpeg}, JPEG2000, and WEBP\cite{shelley2011webp}. The comparison shows that low-rank approximations obtained from Singular Value Decomposition achieve substantially worse compression efficiency across the entire tested range, with especially large gaps at low tolerated error levels. 

\section{Background}

\subsection{Representing Images as Matrices}

A grayscale image can be represented as a two dimensional matrix whose entries correspond to pixel brightness. Colored images can be archived by storing multiple matrices, with each matrix representing red, green, and blue channels, and optionally an alpha channel respectively. Each channel is processed independently, so a colored image may be viewed as a combination of separate grayscale images.\cite{harris2020array}

\subsection{Singular Value Decomposition of a Matrix}\label{AA}
Any matrix of real numbers can be expressed using Singular Value Decomposition, which can be represnted as
\begin{equation}
A = U\Sigma V^\top
\end{equation}
Multiplying two matrices can be interpreted as summing the outer products of columns from the first matrix with rows from the second. \cite{BlumHopcroftKannan2020} This means the decomposition expresses a matrix as
\begin{equation}
    A = \sum_{i} \sigma_i\Sigma_iV^\top_i
\end{equation}

\subsection{Low-Rank Approximation}
Keeping only the first $k$ singular values and the associated singular vectors yields a rank $k$ approximation of the original matrix.\cite{gonzalez2018digital} Because the largest singular values represent the dominant components of the image, the early terms capture the broad structure, while minor details are carried by the smaller singular values. This means truncating the decomposition produces an approximation that preserves major visual information with low rank matrices

\begin{equation}
A_k = \sum_{i = 0}^k \sigma_i\Sigma_iV^\top_i
\end{equation}

\subsection{Quantifying Error of Approximation Image}

The quality of an approximation can be quantified using Frobenius error \cite{malek1983characterization},
\begin{equation}
||A - A_k||F = \sqrt{\sum{i, j} (A_{i, j} - (A_k)_{i, j})^2}.
\end{equation}
Larger images naturally yield larger absolute Frobenius errors because they contain more pixels contributing to the sum. To compare images of different sizes, the relative Frobenius error
\begin{equation} \label{eq:5}
\frac{||A - A_k||_F}{||A||_F}
\end{equation}
is used to normalize the error. This ratio lies between 0.0 and 1.0 and provides a consistent measure across images of varying resolutions \cite{blum2020foundations}.

For Singular Value Decomposition approximations, the Frobenius error also has a closed form expression in terms of the discarded singular values, where the squared Frobenius error equals the sum of the squares of all singular values omitted in the rank $k$ reconstruction.\cite{mirsky1960symmetric} However, this formula only applies for Singular Value Decomposition approximations. Because this study compares against industry standard codecs, whose reconstructions are not represented by truncating singular values, all errors are computed directly from pixel-wise differences using the general definition above.

\subsection{Minimizing Frobenius Error}
The Eckart–Young–Mirsky theorem guarantees that the truncated decomposition minimizes Frobenius error for any rank $k$ approximations. This establishes that using the first $k$ terms of the Singular Value Decomposition produces the best possible approximation under this metric. \cite{schmidt1907theorie}
\subsection{Compression Ratio}
Storage efficiency is evaluated using compression ratio defined as 
\begin{equation}
    1 - \frac{\text{Number of bytes of compressed image}}{\text{Number of bytes of compressed image}}
\end{equation} 

A higher value indicates greater storage savings. Sometimes, the compressed representation may require more space than the original image, producing a negative compression ratio. \cite{hevc2013textspec} 

\section{Methods}
\subsection{Singular Value Decomposition of Images}
Each image is represented as a NumPy array whose dimensions correspond to pixel height and width. Singular Value Decomposition is computed using \verb|np.linalg.svd|, which returns the singular values and the associated left and right singular vectors. The reduced form of the decomposition is used (\verb|full_matrices=False|) to avoid including trailing singular values of 0 and vectors that contribute nothing to the reconstruction. This reduces computation so that the approximation uses only the components that influence the outer product expansion.
\subsection{Computing Relative Frobenius Error}
Relative Frobenius error is computed using \verb|np.linalg.norm|, which calculates the square root of the sum of the squared entries of an array. In this setting, \verb|np.linalg.norm(A - A_k)| computes the Frobenius norm of the pixel wise difference between the approximation and the original image by squaring each pixel difference, summing all squared values, and taking the square root. The relative error is then obtained by dividing this quantity by \verb|np.linalg.norm(A)|, which computes the Frobenius norm of the original image. This implementation directly matches the definition \eqref{eq:5} and describes the approximation quality across images of different sizes.

\subsection{Selecting Tolerated Error Levels}
A set of target relative Frobenius error levels $\{0.05, 0.10, …, 0.95\}$ is used to control approximation quality. For each tolerance, the algorithm increases the rank $k$ of the truncated decomposition until the approximation meets the specified error. Once a suitable rank is reached, the image is reconstructed and encoded to compute its storage size. Compression ratio is then calculated using the encoded size of the approximation and the encoded size of the original image, rather than the sizes of the matrices $U$, $\sigma$, and $V^\top$, to measured the realized gains.

\subsection{Grayscale Experiments}
The procedure is first applied to a grayscale image of dogs, where the image is approximated at all values of $k$. Reconstructed images and pixel-wise error maps are generated for several representative ranks $k=5,10,20,50,100$ to illustrate the effect of rank on reconstruction quality. A graph of compression ratio to relative Frobenius error is also produced.

\subsection{Colored Image Experiments}
Colored images are processed by separating the red, green, and blue channels into independent NumPy arrays. The image is approximated at all values of $k$ for each channel separately, and the approximated channels are then recombined to form the final reconstruction. Reconstructed color image of dogs and channel-wise error maps are generated for the ranks $k=5,10,20,50,100$. A graph of compression ratio to relative Frobenius error is also produced. This procedure can be used on images with any number of channels. This procedure generalizes to images with any number of channels. A demonstration using an image that includes an alpha channel is also provided.

\subsection{Evaluation on ILSVRC2017 Subset}
\begin{figure}[htbp]
\centerline{\includegraphics[width=\columnwidth]{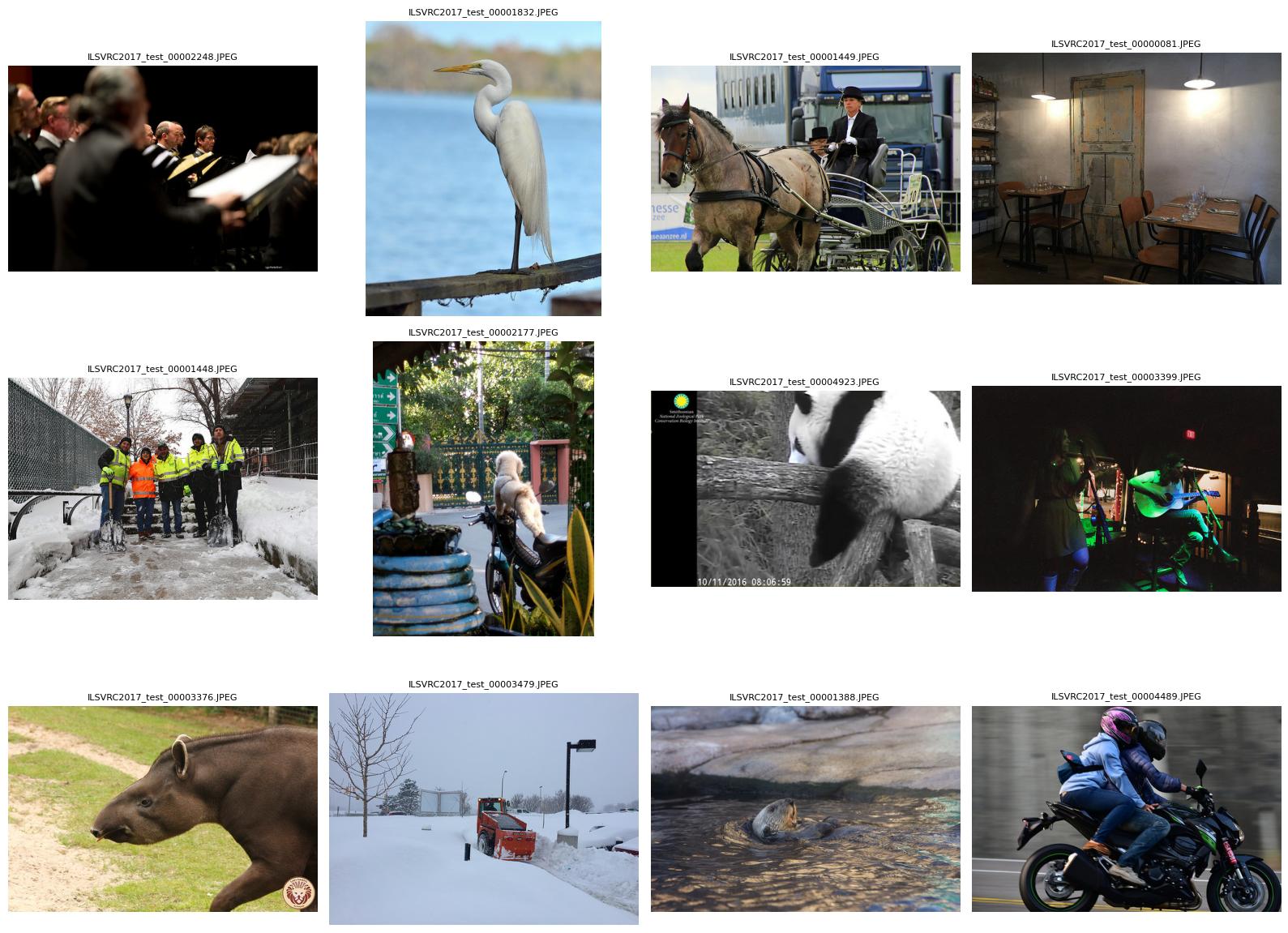}}
\caption{12 random colored pictures of various dimensions from the ILSVRC2017 dataset}
\label{fig:sampleImages}
\end{figure}

To benchmark performance across diverse content, the method is applied to a 1,000 image subset of the ImageNet Large Scale Visual Recognition Challenge 2017 test set.\cite{deng2009imagenet} For each image, the procedure for colored images is repeated and the corresponding compression ratios for each target relative Frobenius error is recorded. A graph of the average compression ratios across all images to the target relative Frobenius errors is also produced. A sample of 12 random pictures of the 1,000 image subset is included in Fig. \ref{fig:sampleImages}

\subsection{Comparison With Industry Codecs}
To contextualize the compression efficiency of using Singular Value Decomposition, the same images are compressed using JPEG, JPEG2000, and WEBP. For these codecs, compression level is controlled by adjusting the quality parameter in \verb|cv2.imencode|. The size of the encoded arrays are used to compute compression ratios for each tolerance level. A graph of the average compression ratios to relative Frobenius errors is also produced for each codec. 

\section{Results}
\subsection{Experiment Results on Grayscale Image}
\begin{figure}[htbp]
\centerline{\includegraphics[width=\columnwidth]{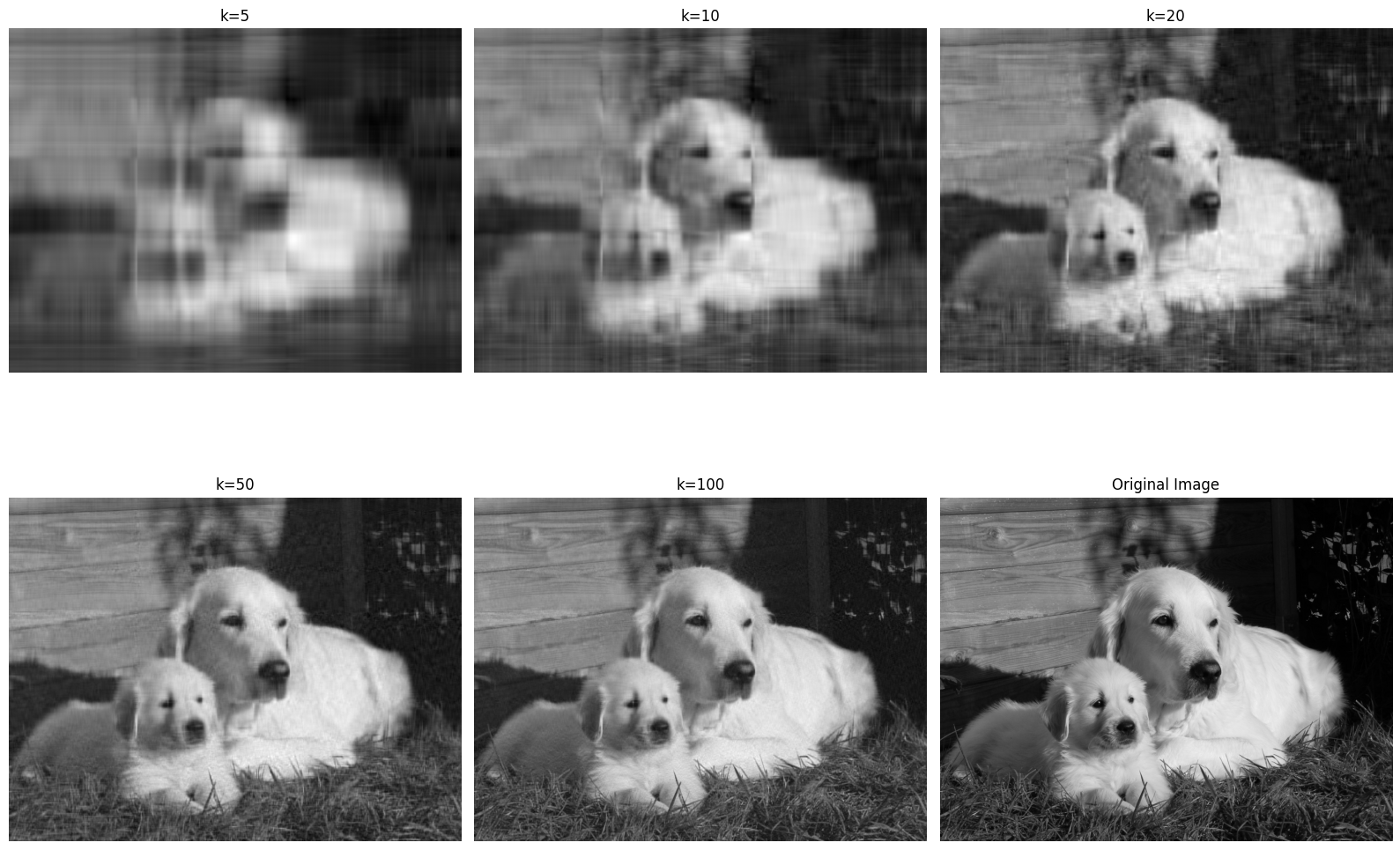}}
\caption{Grayscale image of dogs using rank $k = 5, 10, 20, 50, 100$ approximation (singular value decomposition) and original grayscale image.}
\label{fig:GrayKs}
\end{figure}

\begin{figure}[htbp]
\centerline{\includegraphics[width=\columnwidth]{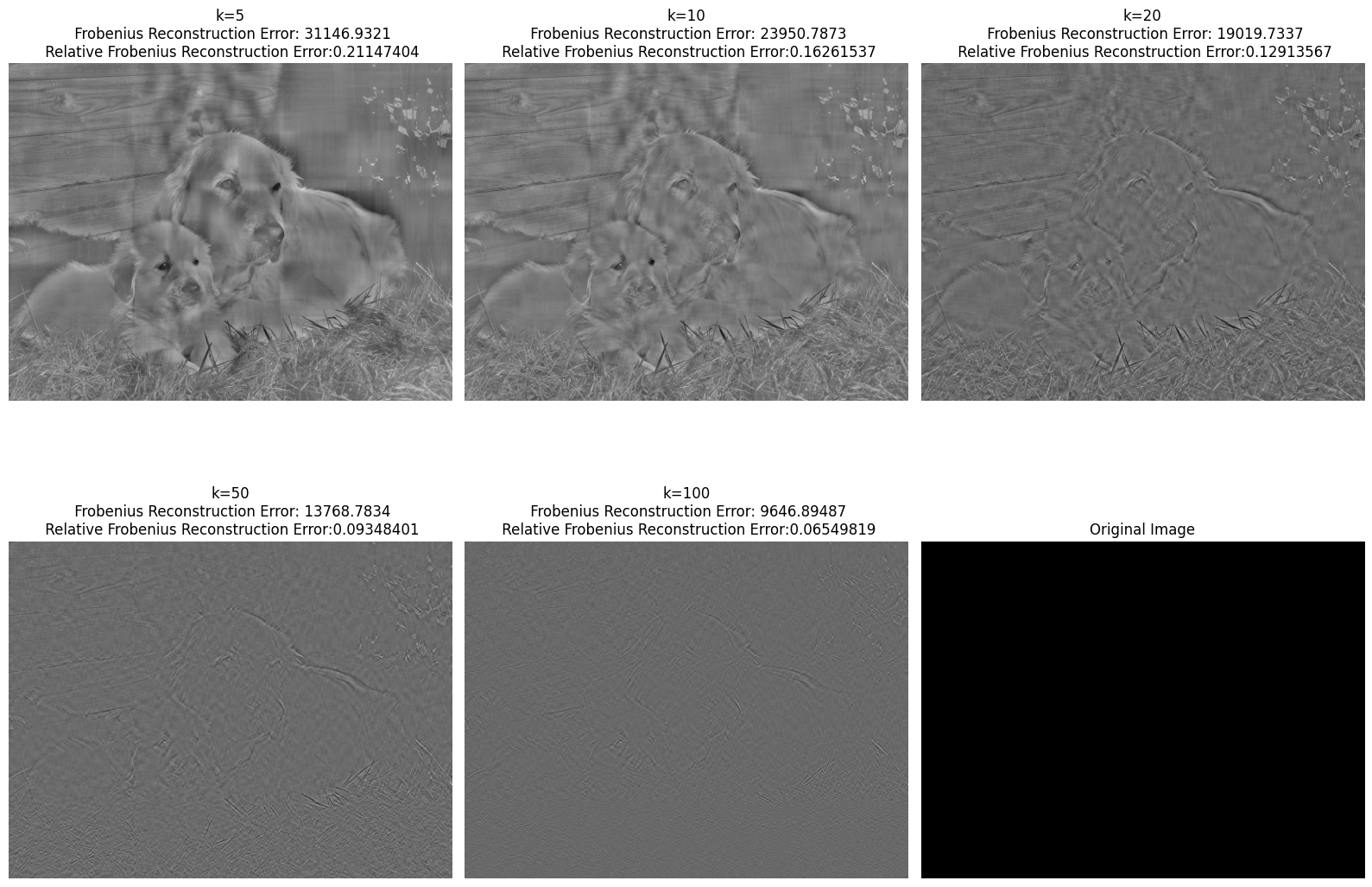}}
\caption{Visualization of error of approximation using rank $k = 5, 10, 20, 50, 100$ approximation(singular value decomposition) and original grayscale image. Brighter areas signify higher pixel wise error.}
\label{fig:GrayError}
\end{figure}

\begin{figure}[htbp]
\centerline{\includegraphics[width=\columnwidth]{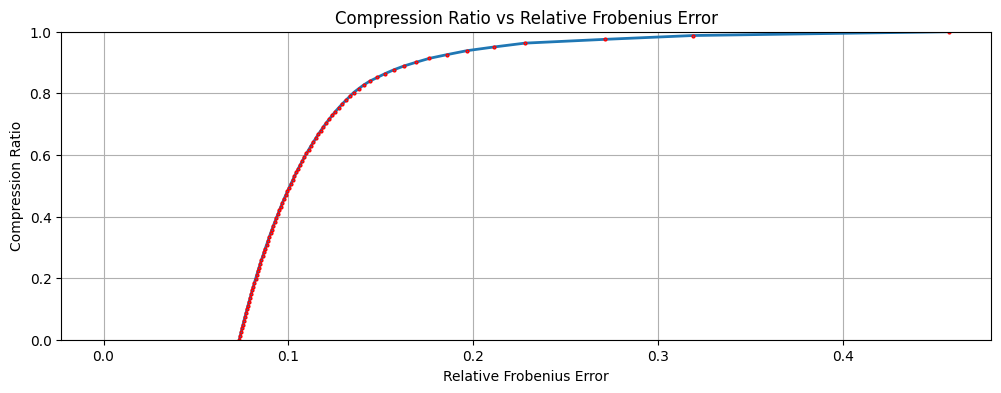}}
\caption{Compression Ratio vs Relative Frobenius Error for grayscale Image}
\label{fig:GrayCRvsRFE}
\end{figure}

Reconstructed grayscale images for several ranks ($k=5,10,20,50,100$) using singular value decomposition are shown in Fig. \ref{fig:GrayKs} along with the original image. The corresponding pixel-wise error maps are presented in Fig. \ref{fig:GrayError}. A curve relating compression ratio to relative Frobenius error for the grayscale image is provided in Fig. \ref{fig:GrayCRvsRFE}.

\subsection{Experiment Results on Colored Image}
\begin{figure}[htbp]
\centerline{\includegraphics[width=\columnwidth]{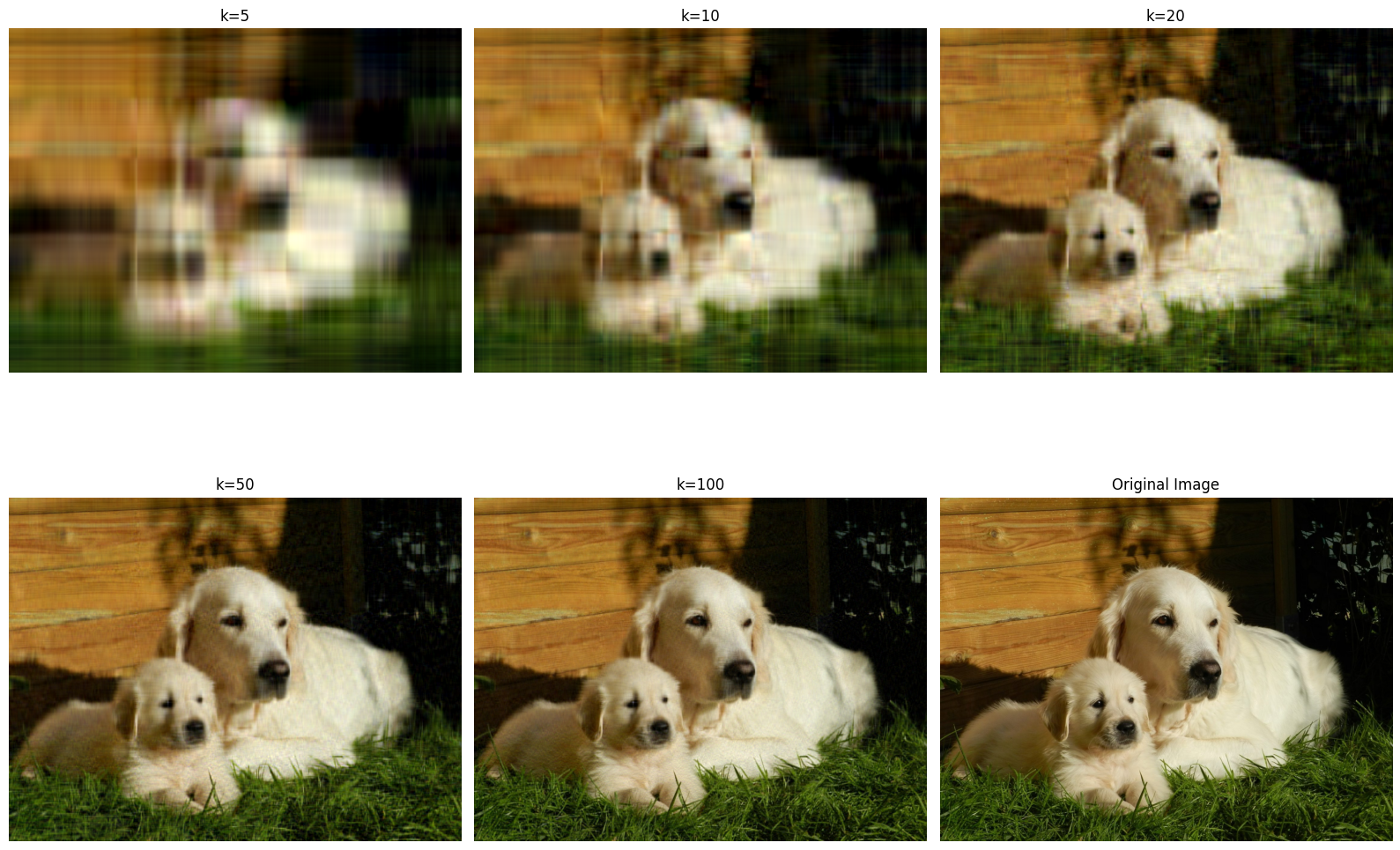}}
\caption{Colored image of dogs using rank $k = 5, 10, 20, 50, 100$ approximation (singular value decomposition) and original colored image.}
\label{fig:ColoredKs}
\end{figure}

\begin{figure}[htbp]
\centerline{\includegraphics[width=\columnwidth]{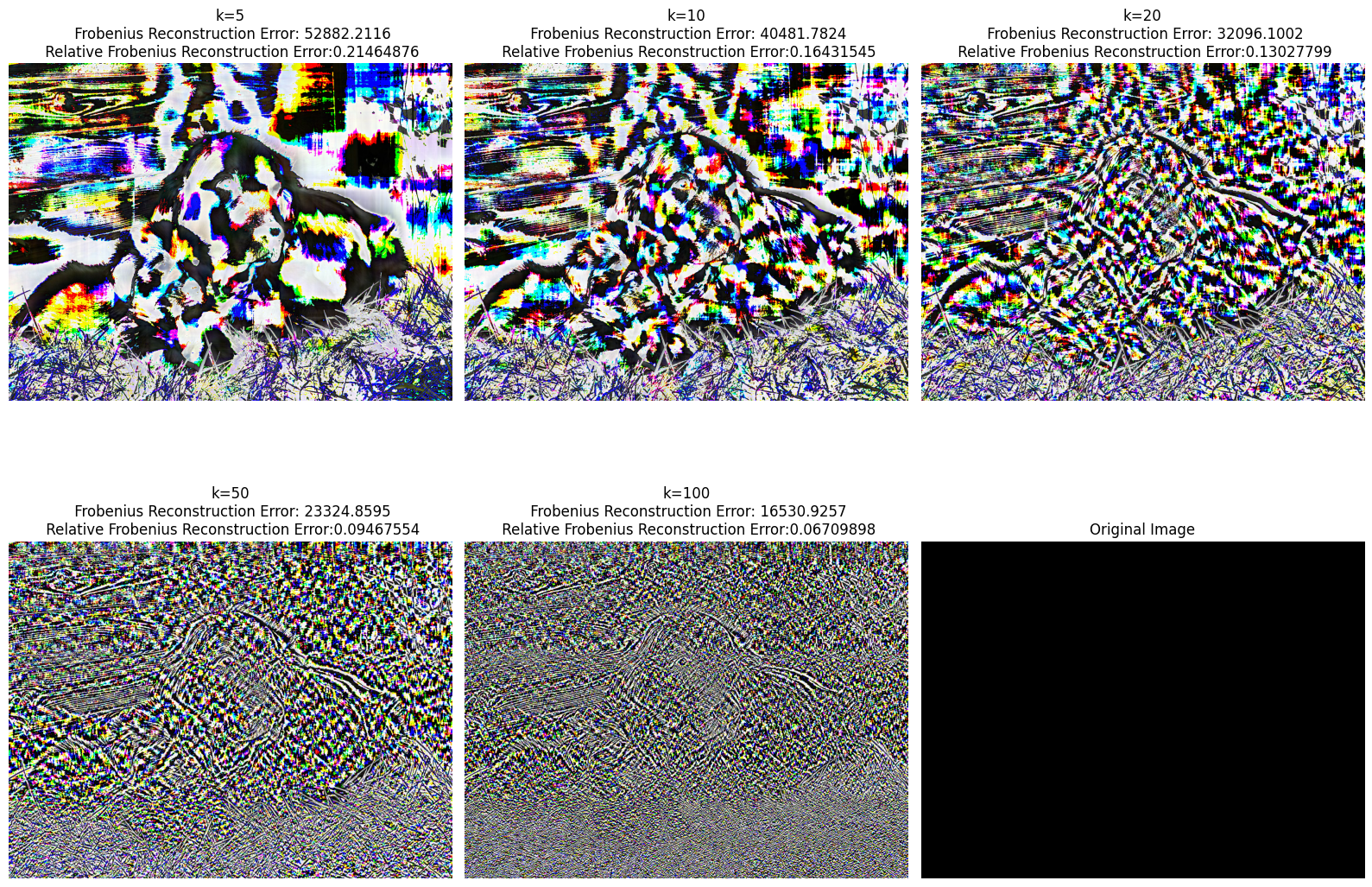}}
\caption{Visualization of error of approximation using rank $k = 5, 10, 20, 50, 100$ approximation(singular value decomposition) and original colored image. Brighter areas signify higher pixel wise error.}
\label{fig:ColoredError}
\end{figure}

\begin{figure}[htbp]
\centerline{\includegraphics[width=\columnwidth]{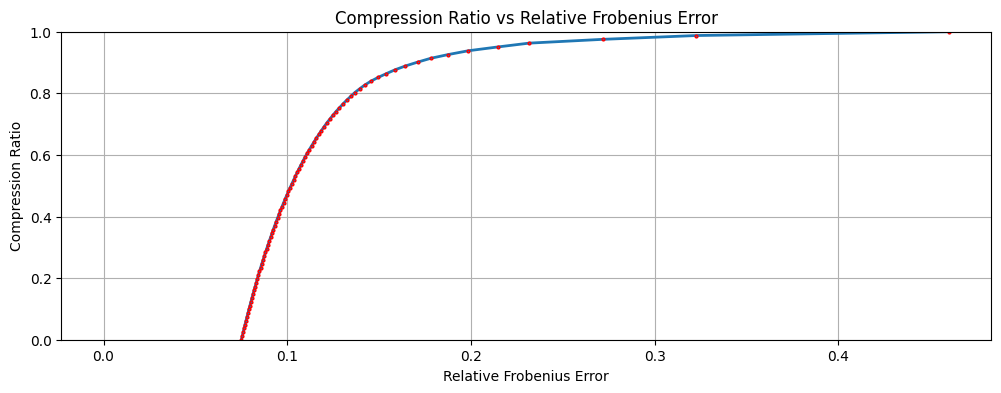}}
\caption{Compression Ratio vs Relative Frobenius Error for colored Image}
\label{fig:ColoredCRvsRFE}
\end{figure}

\begin{figure}[htbp]
\centerline{\includegraphics[width=\columnwidth]{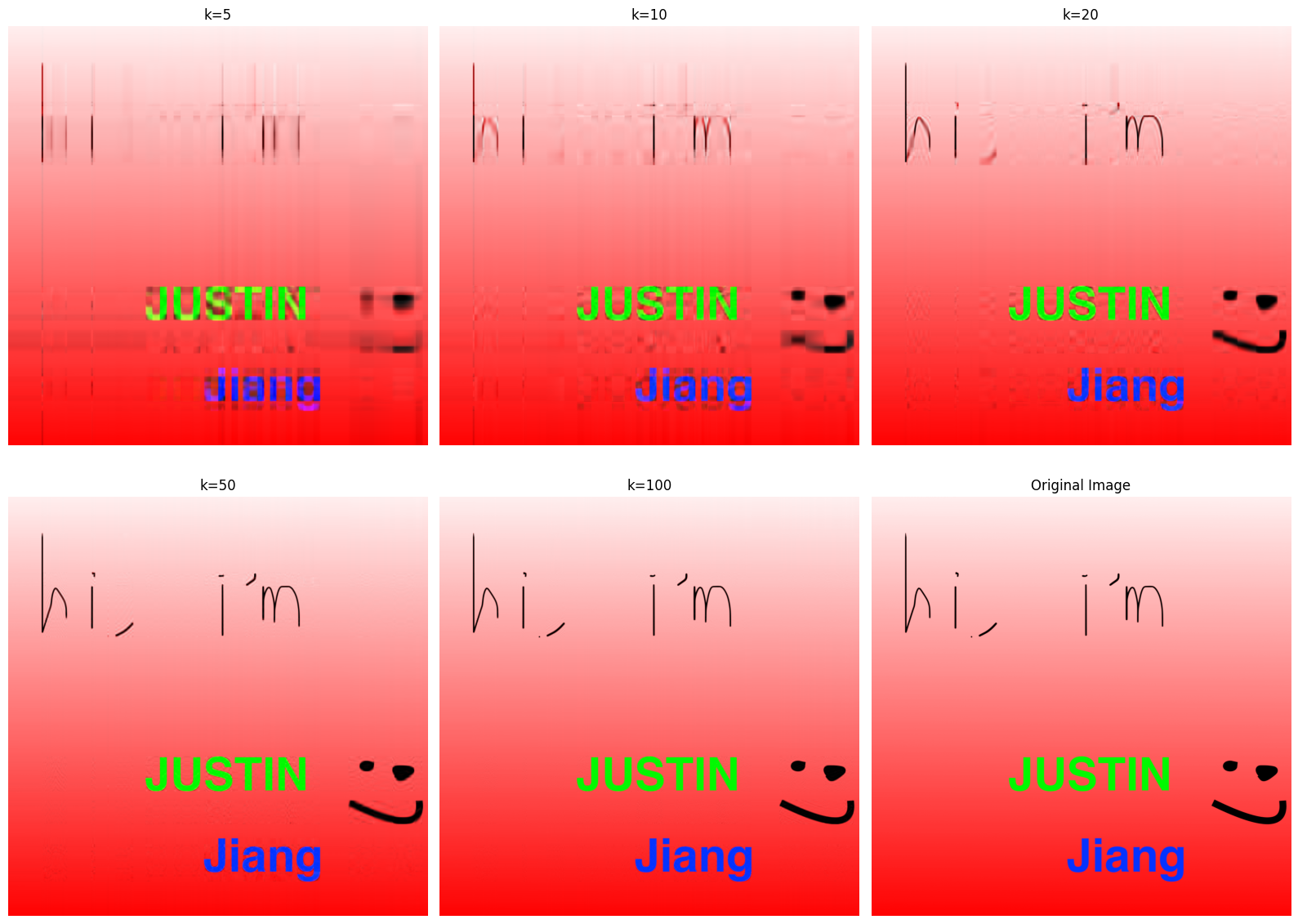}}
\caption{Colored image containing alpha channel using rank $k = 5, 10, 20, 50, 100$ approximation (singular value decomposition) and original colored image.}
\label{fig:RedAlpha}
\end{figure}

Reconstructed color images for the selected ranks are shown in Fig. \ref{fig:ColoredKs}. Pixel wise error visualization is displayed in Fig. \ref{fig:ColoredError}. The compression ratio versus relative Frobenius error for the colored image is shown in Fig. \ref{fig:ColoredCRvsRFE}. Reconstructed color images with alpha channel for the selected ranks are shown in Fig. \ref{fig:RedAlpha}.

\subsection{Industry Codecs and Singular Value Decomposition on ILSVRC2017}
\begin{figure}[htbp]
\centerline{\includegraphics[width=\columnwidth]{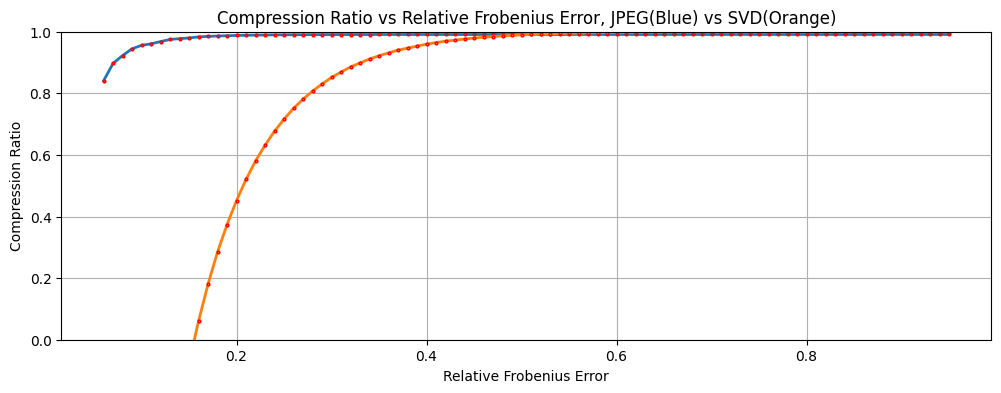}}
\caption{Compression Ratio vs Relative Frobenius Error Graph for JPEG Codec(Blue line) and singular value decomposition approximation(Orange line)}
\label{fig:SVDvsJPEG}
\end{figure}

\begin{figure}[htbp]
\centerline{\includegraphics[width=\columnwidth]{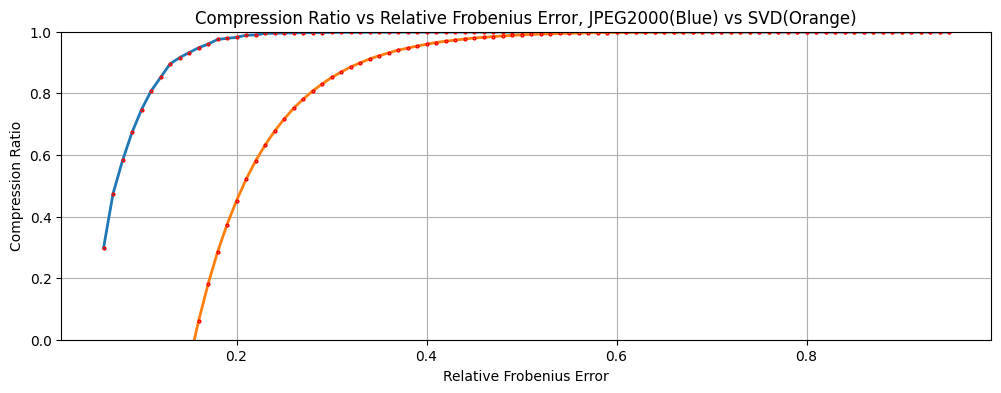}}
\caption{Compression Ratio vs Relative Frobenius Error Graph for JPEG2000 Codec(Blue line) and singular value decomposition approximation(Orange line)}
\label{fig:SVDvsJPEG2000}
\end{figure}

\begin{figure}[htbp]
\centerline{\includegraphics[width=\columnwidth]{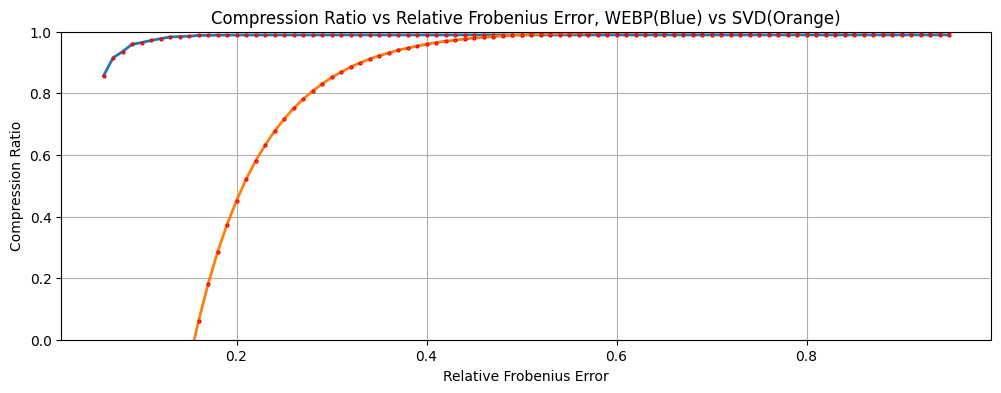}}
\caption{Compression Ratio vs Relative Frobenius Error Graph for WEBP Codec(Blue line) and singular value decomposition approximation(Orange line)}
\label{fig:SVDvsWEBP}
\end{figure}

Across the 1,000-image subset, compression ratio is evaluated at matched relative Frobenius error levels for Singular Value Decomposition, JPEG, JPEG2000, and WEBP. Figures \ref{fig:SVDvsJPEG}–\ref{fig:SVDvsWEBP} show the corresponding curves for each comparison.

\section{Discussion}
\subsection{Areas of High Pixel-Wise Error}
The visualization of erros for the grayscale image show that low rank approximations have concentrated error around object boundaries, such as the edges between the dog and the background or between the grass and the dogs. These smaller details are represented by the smaller singular values, which are lost in the approximation. As the rank increases, this error diminishes, and the residuals become more diffuse and less correlated with the structures in the original image. This same pattern appears in the colored image, where lower rank approximations have noticeable structural artifacts that resemble the boundaries in the original image, while higher ranks reduce these patterns and distribute the remaining error more evenly.

\subsection{Non Linear Relation Between Compression Ratio And Error Tolerance}
Both the grayscale and colored experiments show a highly nonlinear relationship between compression ratio and tolerated relative Frobenius error. Small increases in tolerated error near zero yield negative compression ratios, where storing the ``low rank'' matrices for the approximation takes more storage than storing the original image. Once the tolerance increases beyond this regime, the compression ratio improves sharply. For larger tolerated errors, the gain in compression ratio gradually stabilizes, indicating diminishing returns. These results imply that achieving very low relative Frobenius error requires retaining a large number of singular values, dramatically reducing compression efficiency, whereas moderate tolerances allow for substantially better compression.

\subsection{Singular Value Composition Approximation And Industry Standard Codecs}
The same trend appears when comparing industry codecs with the Singular Value Decomposition approximation, all codecs achieve poor compression ratios when the tolerated error is small due to the cost of maintaining the granular details. However, the degradation is significantly more severe for Singular Value Decomposition. Unlike JPEG, JPEG2000, and WEBP, which use quantization, entropy coding, block or wavelet decompositions, and perceptual optimizations to reduce storage\cite{7924246}, Singular Value Decomposition stores large dense matrices directly. 

Compression ratio versus relative Frobenius error graphs also show that Singular Value Decomposition reaches compression efficiencies similar to JPEG, JPEG2000, and WEBP only when the tolerated relative Frobenius error is very high. In this regime, all methods achieve near-maximal compression ratios, and the differences between codecs are small. As the tolerated error decreases, however, the gap widens rapidly. At approximately 50\% relative Frobenius error, the compression ratios produced by Singular Value Decomposition and the industry codecs differ by about 1\%. At approximately 80\% relative Frobenius error, the compression ratio for Singular Value Decomposition is roughly half that of the other formats. These results indicate that when large errors are permitted, the choice of codec has little influence on compression ratio, but at lower tolerated errors, the compression efficiency of Singular Value Decomposition is significantly worse.

\section{Conclusion}
The experiments in this study shows that Singular Value Decomposition–based low-rank approximation can reproduce many visual features of an image, but its storage efficiency is highly sensitive to the tolerated reconstruction error. When low relative Frobenius error is required, the number of retained singular values grows quickly, and the resulting compression ratio becomes low. At higher tolerated errors, the method achieves noticeably better compression, and the differences across codecs narrow.

In comparisons with JPEG, JPEG2000, and WEBP, the Singular Value Decomposition approach consistently produced lower compression ratios across the range of error tolerances tested. The findings suggest that Singular Value Decomposition approximations are mostly useful for academic purposes, it does not offer competitive compression performance relative to industry standard image formats.

Future work could explore incorporating quantization into the low-rank representation, evaluating the effect of alternative color spaces on approximation quality.

\bibliographystyle{ieee}
\bibliography{citations}
\end{document}